\documentclass[a4paper]{article}

\usepackage{INTERSPEECH2022}
\usepackage{multirow}
\usepackage[table,xcdraw]{xcolor}
\usepackage{hyperref}
\usepackage{enumitem}

\title{BERT, can HE predict contrastive focus? Predicting and controlling prominence in neural TTS using a language model}
\name{Brooke Stephenson$^{1,2}$, Laurent Besacier$^{2,3}$, Laurent Girin$
{^1}$, Thomas Hueber${^1}$}
\address{
  $^1$Université Grenoble Alpes, CNRS, Grenoble INP, GIPSA-lab, 38000 Grenoble, France\\
  $^2$LIG, UGA, G-INP, CNRS, INRIA, Grenoble, France\\
  $^3$NAVER LABS Europe, Meylan, France}
\email{brooke.stephenson@grenoble-inp.fr, 
thomas.hueber@grenoble-inp.fr,
laurent.girin@grenoble-inp.fr,
laurent.besacier@univ-grenoble-alpes.fr}

\begin{document}

\maketitle
\begin{abstract}
  Several recent studies have tested the use of transformer language model representations to infer prosodic features for text-to-speech synthesis (TTS). While these studies have explored prosody in general, in this work, we look specifically at the prediction of contrastive focus on personal pronouns. This is a particularly challenging task as it often requires semantic, discursive and/or pragmatic knowledge to predict correctly. We collect a corpus of utterances containing contrastive focus and we evaluate the accuracy of a BERT model, finetuned to predict quantized acoustic prominence features, on these samples. We also investigate how past utterances can provide relevant information for this prediction. Furthermore, we evaluate the controllability of pronoun prominence in a TTS model conditioned on acoustic prominence features.

\end{abstract}

\section{Introduction}


\begin{figure*}[!]
  \centering
  \includegraphics[width=0.7\linewidth]{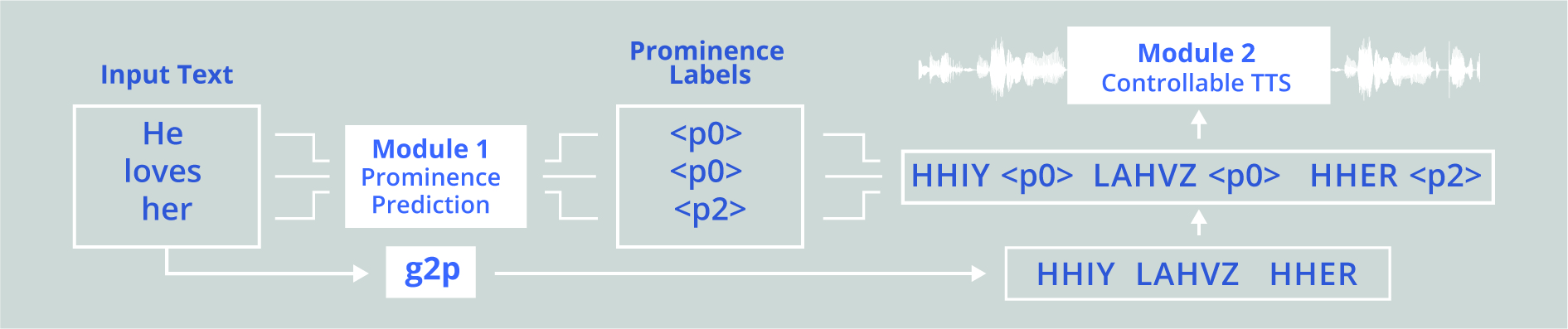}
  \caption{TTS overview: The system is split into two modules. The first uses a language model to predict prominence labels. The second controls the prominence in the synthetic speech in accordance with the predicted labels.}
  \label{fig:overview}
\end{figure*}

``\textit{HE loves her}'', ``\textit{he LOVES her}'', and ``\textit{he loves HER}'' all have the same textual content, but three distinct communicative goals. Indeed, such \textit{contrastive focus} is used by speakers to evoke alternative sets in the discourse \cite{rooth1992theory}. This can be utilized to make explicit intended discourse relations between clauses/paragraphs/sections, to highlight a fact that the listener may find surprising, or to express a specific semantic or pragmatic meaning. 
The prediction of contrastive focus placement therefore often requires high-level linguistic understanding. Current vanilla neural text-to-speech (TTS) synthesis systems lack this understanding and will always pronounce the above sentences in the same way, irrespective of the context. In this work, we investigate methods to predict the placement of contrastive focus and to control it in a TTS system. Figure \ref{fig:overview} illustrates our overall approach which addresses both \textit{predicting} and \textit{controlling} prominence.



One way to insert sophisticated linguistic information may be through the use of contextualized word embeddings. While other works have explored the use of transformer language models to predict prosodic and stylistic features \cite{Kenter2020ImprovingTP, zou2021fine, contextEmbProm, 9054337, hodari2021camp, stanton2018predicting, hayashi19_interspeech}, it has not been fully explored how much the encoded word representations actually imbue high-level knowledge. In other words, do they provide information about the content and context of the message or do they only provide/reinforce low-level linguistic features such as the likelihood of lexical prominence, parts of speech and position in the sentence? For prominence/pitch accent prediction, a fairly high baseline can be achieved using word majority/accent-ratio alone (i.e., if a lexical item is usually prominent in the training set, it is likely to be prominent in the test set) \cite{nenkova-etal-2007-memorize}. Moreover, \cite{Hirschberg95pitchaccent} found that in a binary pitch accent prediction task, using broad word category distinctions (open/content or closed/function) could achieve 68\% accuracy; more fine-grained division of the closed class category brought that number up to 77\%. In this work, we probe a language model, in the present case BERT \cite{devlin-etal-2019-bert}, by choosing a testing ground that cannot rely on simple heuristics to achieve good results: the prediction of contrastive focus on personal pronouns. Personal pronouns, at least in the corpora typically used to train TTS systems, are majority non-prominent.\footnote{We conflate the notions of prominence and contrastive focus for personal pronouns; when they are prominent, they typically possess a contrastive meaning.}   

Contrastive focus in English is mainly realized prosodically through increases in pitch, energy and/or duration. Suni et al.~\cite{Hierarchical} have proposed extracting prosodic information automatically through a combined representation of these signals. They use continuous wavelet transforms, which analyze the speech signal at different timescales, to identify acoustically prominent words. In \cite {contextEmbProm}, they tested the prediction capabilities of a language model on these features. Here, we extend this investigation by looking at a challenging set of contrastively focused pronouns and by looking at the use of an extended context (current sentence, +previous sentence, +2 previous sentences). Moreover, because the \textit{prediction} of contrastive focus is complicated by the one-to-many issue that hinders TTS evaluation in general, we have collected a corpus of audiobooks where we have three separate renditions of each book, each read by a separate speaker. We use these three realizations to study consensus among speakers in the placement of contrastive focus.

Assuming we are able to predict good candidates for contrastive focus, it remains to be verified if we can \textit{control} the contrastive prominence on pronouns in a neural TTS system.  Other works have proposed utterance-level control systems \cite{controllable-neural-text-to-speech-synthesis,wang2018style} but for this task we need to target the word-level.  This was achieved previously within an HMM framework \cite{badino2009identification}. 
\cite{zou2021fine} evaluated the expression of prosodic features from ToBI-label conditioned TTS, but only for ground-truth input labels.  
\cite{conditionedOnProm} report successful control over other word categories from systems conditioned on human or automatically annotated data (although this was not evaluated with a perceptual test). More recently, \cite{latif-etal-2021-controlling} proposed to use control tags for prominence in an end-to-end TTS system. However, their system relied on the availability of a specific (and limited) pre-annotated corpus whereas we demonstrate controllability from prominence labels obtained automatically.

\vspace{10mm}

\begin{table*}[t]
 \caption{Results for the prominence prediction task for the $<$p2$>$ (high prominence) category. Recall (R), Precision (P) and F1 are reported for the full test set (all POS categories combined). }
 \label{tab:predictionResults}
\begin{tabular}{llllllllllll}
\hline
\textbf{Model}  & \multicolumn{3}{l}{\textbf{Current sentence}} & \multicolumn{3}{l}{\textbf{+previous sentence}} & \multicolumn{3}{l}{\textbf{+2 previous sentences}} & \textbf{\# p2 tags} & \textbf{\# words} \\ 
 \multicolumn{1}{l}{} & \textbf{\footnotesize F1 } & \textbf{\footnotesize R } & \textbf{\footnotesize P } & \textbf{\footnotesize F1 } & \textbf{\footnotesize R } & \textbf{\footnotesize P } & \textbf{\footnotesize F1 } & \textbf{\footnotesize R } & \textbf{\footnotesize P } &  &  \\ 
 \hline
 \multicolumn{1}{l|}{\footnotesize Word majority} & \footnotesize 0.458 & \footnotesize 0.394 & \footnotesize 0.546 & \multicolumn{6}{l}{\cellcolor[HTML]{EFEFEF}}  & \footnotesize 79,793 & \footnotesize 365,330 \\
 \multicolumn{1}{l|}{\footnotesize Randomly initialized BERT}  & \footnotesize 0.572 & \footnotesize 0.539 & \footnotesize 0.609 & \footnotesize 0.565 & \footnotesize 0.546 & \footnotesize 0.586 & \footnotesize 0.561 & \footnotesize 0.558 & \footnotesize 0.565 & \footnotesize 79,793 & \footnotesize 365,330 \\
 \multicolumn{1}{l|}{\footnotesize Fine-tuned BERT}  & \footnotesize 0.588 & \footnotesize 0.552 & \footnotesize 0.629 & \footnotesize 0.580 & \footnotesize 0.536 & \footnotesize 0.632 & \footnotesize 0.579 & \footnotesize 0.535 & \footnotesize 0.629 & \footnotesize 79,793 & \footnotesize 365,330 \\
 \hline
\end{tabular}
\end{table*}

\begin{table}[]
 \caption{Results on pronouns only for the prominence prediction task. Recall is reported for the two manually verified subsets of contrastively focused personal pronouns and the non-contrastive pronoun subset (Neg: 493 samples). Maj: group of 393 samples where majority of speakers used contrastive focus. Min: group of 100 samples where only 1 out of 3 speakers used contrastive focus. Context: current (Curr), previous (Prev)}
 \label{tab:pronouns}
\begin{tabular}{l|c|c|c|c|c}
\hline
\small \textbf{Model}                     & \small \textbf{Data} & \small \textbf{Cat}. & \small \textbf{R} & \small \textbf{R}  & \small \textbf{R}  \\ \hline
\textbf{Context} & & & \textbf{Curr} & \textbf{+1 Prev} & \textbf{+2 Prev}
\\ \hline
\footnotesize Word             & \scriptsize Maj        & \scriptsize $<$p2$>$ & \scriptsize 0.079               &                        &                      \\
 \footnotesize Majority               & \scriptsize Min    & \scriptsize $<$p2$>$    & \scriptsize 0.000               &                        &                      \\ 
                           & \scriptsize Neg    & \scriptsize $<$p0$>$     & \scriptsize 1.000               &                        &                      \\ 
                          \hline
\scriptsize \footnotesize Randomly & \scriptsize Maj  & \scriptsize $<$p2$>$      & \scriptsize 0.178               & \scriptsize 0.160 & \footnotesize 0.135                \\
       Initialized    & \scriptsize Min  & \scriptsize $<$p2$>$      & \scriptsize 0.060               & \scriptsize 0.060                  & \scriptsize 0.060                \\
        \footnotesize BERT          & \scriptsize Neg  & \scriptsize $<$p0$>$      & \scriptsize 0.980               & \scriptsize 0.980                  & \scriptsize 0.982                \\
                          \hline
\footnotesize Fine-   & \scriptsize Maj  & \scriptsize $<$p2$>$      & \scriptsize 0.239               & \scriptsize 0.239                  & \scriptsize 0.216                \\
\footnotesize tuned              & \scriptsize Min   & \scriptsize $<$p2$>$     & \scriptsize 0.060               & \scriptsize 0.070                  & \scriptsize 0.070                \\ 
\footnotesize BERT              & \scriptsize Neg   & \scriptsize $<$p0$>$     & \scriptsize 0.990               & \scriptsize 0.986                  & \scriptsize 0.988                \\ 
                          
                          \hline
\end{tabular}
\end{table}

\section{Prepared Datasets}
\label{ccreation}

The selected audiobooks for our corpus are literary texts sourced from Librivox\footnote{\scriptsize \url{https://librivox.org}} and from the Blizzard Challenge 2013 dataset.\footnote{\scriptsize \url{https://www.synsig.org/index.php/Blizzard\_Challenge\_2013}}. Criteria for book selection included open-source status, the availability of multiple recordings with different speakers (minimum 3), audio quality and the subject matter (we favoured books dealing with interpersonal relationships as they are more likely to contain contrastively focused pronouns). Five novels (x 3 speakers) were selected for the training set (41,593 utterances, approx. 66 hours of audio/speaker\footnote{Two of the five book sets are single speaker. The third set is made up of multiple speakers.} and one novel (x 3 speakers) was selected for testing (6838 utterances, approx. 11 hours of audio per speaker). The corresponding transcriptions were obtained from Project Gutenberg.\footnote{\scriptsize \url{https://www.gutenberg.org}} Transcripts were split into chapters and then sentences using \cite{chapterize} and \cite{Honnibal_spaCy_Industrial-strength_Natural_2020}. The audio files were segmented into utterances using the Aeneas library\footnote{\scriptsize \url{https://github.com/readbeyond/aeneas}} and phoneme alignment was obtained using the Montreal Forced Aligner \cite{mcauliffe17_interspeech}.  \\
\textbf{Prosodic feature extraction}. The audio files were analyzed with the continuous wavelet transform (CWT) method of \cite{Hierarchical} implemented in the Wavelet Prosody Toolkit.\footnote{\scriptsize \url{https://github.com/asuni/wavelet_prosody_toolkit}} This method assigns a prominence score to each word in the corpus. This is done by combining f0, energy and duration into a composite signal, performing the CWT, establishing lines of maximum amplitude connecting the various timescales and then calculating a weighted sum of the points in this line (see \cite{Hierarchical} for details). We then quantize this score into three categories: $<$p2$>$ (strong prominence), $<$p1$>$ (intermediate prominence) and $<$p0$>$ (no prominence).\footnote{Dataset available at https://doi.org/10.5281/zenodo.6646827.}

\noindent \textbf{Contrastive personal pronoun subcorpus}. With the processed data from the previous section, we searched for utterances containing $<$p2$>$ labelled personal pronouns (strong prominence) in the test set. With manual verification, we collected positive examples of contrastive focus on pronouns. We randomly selected an equal number of negative samples where the pronoun was tagged as $<$p0$>$ for all three speakers; these samples were also manually checked. We then enlisted three native English speakers to validate 200 pronouns (x 3 speakers) from the collected samples (100 positive samples and 100 negative samples, just over 20\% of the full pronoun corpus). Validators were presented with the audio clips (for all three speakers) and a transcription of the text with the pronoun of interest highlighted in red. They were asked to assign a value of 1 if they deemed the speaker had used prosody to convey contrastive focus and 0 if they did not. Cohen-kappa scores \cite{cohen} were used to evaluate inter-annotator agreement between the three raters. These scores range from 0.85--0.90; this shows strong to almost perfect agreement.  
For evaluation purposes, we sorted the positive samples into two groups: 1) those where the majority of speakers (at least 2 out of 3) used contrastive focus (\textbf{Pronoun maj.}). This group contains 393 pronouns; 310 of which are prosodically contrasted by all three speakers; 2) those where only 1 out of 3 speakers used contrastive focus (\textbf{Pronoun min.}). This group contains 100 pronouns. All contrastive pronouns come from 406 utterances as several utterances contain multiple examples. 

In this study, our intention is to find challenging examples for prominence prediction (i.e., words that are not frequently prominent). Subjective pronouns (e.g., I, we), objective pronouns (e.g., me, us) and possessive determiners (e.g., my, our) all fit this requirement, but possessive pronouns (e.g., mine, ours) are more often prominent than not. We decided to include these ``easy'' words in the corpus because they may be of interest for future work on contrastive focus. However, they do not present a particular challenge to prominence prediction.

\section{Predicting Prominence} \label{predicting}

\begin{figure}[t]
  \centering
  \includegraphics[width=\linewidth]{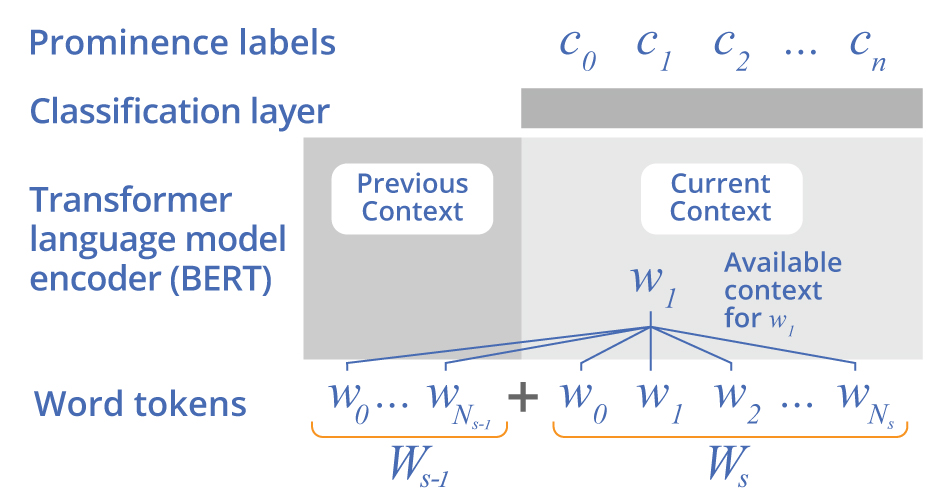}
  \caption{Module 1: Predicting prominence}
  \label{fig:module1}
\end{figure}

\textbf{BERT}. For our prominence prediction task, we used English BERT \cite{devlin-etal-2019-bert}, available on HuggingFace.\footnote{\scriptsize \url{https://huggingface.co/bert-base-cased}} BERT is a transformer encoder that is trained on a masked language modeling task (i.e., it learns to predict masked words using information from the other words in the sentence). Bert makes use of self-attention layers and positional embeddings to form contextualized representations of words.  

\noindent \textbf{Prediction task}.
For the sequence of words $\{w_0, w_1,...,w_N\}$ our objective is to predict a sequence of prominence labels $\{c_0, c_1,...,c_N\}$, where $c_n$ is either $<$p0$>$, $<$p1$>$ or $<$p2$>$. Since knowledge about the previous context is sometimes essential for determining whether a word should be contrastive or not, we experiment with models conditioned on different degrees of past context. 
We note $W_s = \{w_1,\ldots,w_{N_s}\}$ the sequence of $N_s$ words in the current sentence $s$ which we want to synthesize. 
The BERT model is given either the current sentence only, i.e.,~$W_s$, or both the previous and the current sentences, i.e.,~$\{W_{s-1},W_s\}$, or both the two previous sentences and the current one, i.e.,~$\{W_{s-2},W_{s-1},W_s\}$. 
See Figure \ref{fig:module1} for an illustration of the model's components.

\noindent \textbf{Models and linguistic knowledge.}
We evaluate three methods for prominence prediction with increasing access to linguistic knowledge:
\begin{itemize}[leftmargin=*]
    \item Our baseline is a simple word majority method: word statistics from the training corpus are computed; we count how often a lexical item belongs to each of the three prominence categories and the majority category is used for all predictions in the test set. 
    \item The second method involves the use of the BERT architecture, but instead of using weights pretrained on a masked language modelling task, we randomly initialize the model and train it to predict the prominence labels for each word in the input sequence. This model can presumably learn the same word statistics available to the word majority method and additionally the self-attention layers and positional embeddings provide the model with information about the surrounding lexical items and the positional context of each word. We expect this model will be able to learn canonical patterns of English prosody (i.e.~that prominent, nuclear accents are typically found at the end of an intonational phrase) even if the semantic knowledge about the content of the sentences will be non-existent or at best, very naive. 
    \item The third method involves finetuning a pretrained BERT model on the prominence features (learning rate $=5\mathrm{e}{-5}$). From the beginning of training, this model has access to the syntactic and semantic representations learned from training on huge amounts of textual data; during the finetuning process it must find the optimal way to use this information for the prosody/prominence prediction task.
\end{itemize}

\noindent \textbf{Results}. The results for the full dataset are shown in Table~\ref{tab:predictionResults} and the results for the pronoun subsets are shown in Table \ref{tab:pronouns}.
Analyzing these results, we notice that  performance on the contrastive pronoun sets is significantly lower than the full dataset (Recall with finetuned BERT is 0.239 for the pronoun majority group and 0.552 for all POS). The tokens in the pronoun minority group are very rarely predicted to be prominent (highest recall score$=0.07$). Furthermore, we see that BERT does provide some improved prediction accuracy over the two baselines, but the improvement is fairly small.  Word majority method correctly predicts possessives (e.g., mine, yours), and the randomly initialized BERT learns structurally/positionally prominent positions; it correctly predicts prominence at the ends of prosodic phrases (immediately preceding punctuation marks) and following the word `for' (e.g., ``For YOU are...''). We performed McNemar’s  Test \cite{mcnemar1947note} to compare models and we find significant differences (p-values $<0.05$) between each of the three models. With regards to the use of previous context, we do not see any improvements in prediction performance (p-values $>0.05$).

We can imagine several possible causes for low prediction scores. It may be that we have an insufficient number of samples of contrastively focused pronouns to train the model to recognize focus patterns; there is an average of 7893.6 $<$p2$>$ labelled personal pronouns/speaker in the training set, and given the complexity of the task, this may not be enough. 
Alternatively,  learnt representations may not be sophisticated enough to encode the higher level linguistic information required for this task. The recent research trend in language modeling is to scale models bigger and bigger and this increase in size results in better quality on tasks such as text generation. In future work, we will explore using larger LMs.
Finally, as can be expected with any automatic annotation method, there is some noise in the data: 
we did find examples for which prominence was questionable, predominately at phrase boundaries (tagged $<$p2$>$ because of a sharp rise in f0). Hence, human intervention may still be necessary for better fine-grained annotation/control of prosodic data.

\section{Controlling Prominence}

%

\noindent \textbf{Controllable TTS Model}. To synthesize speech with controllable prominence, we follow the method proposed in \cite{conditionedOnProm} where the TTS is conditioned on prominence labels. Our implementation differs in that we use FastSpeech 2 \cite{ren2020fastspeech}
(as implemented by \cite{wang2021fairseqs2}) instead of DC-TTS and we refrain from using boundary tags as we are primarily interested in prominence control. FastSpeech 2 is a non-autoregressive, transformer encoder-decoder model that makes explicit duration, f0 and energy predictions between the encoder and decoder.
The input to our model is a sequence of phonemes and prominence labels. 
Each word $w_n$ in the utterance is converted into a sequence of phonemes $\{p_{n,0},...,p_{n,m}\}$ and this phoneme sequence is followed by the prominence label $c_n$ for $w_n$ (phonemes and prominence labels are converted into embeddings in the first layer of the model).  The output of the modified FastSpeech 2 model is a Mel-spectrogram, which is converted into a waveform using a Parallel WaveGAN vocoder \cite{parallelWavGAN}.\footnote{\scriptsize \url{https://github.com/kan-bayashi/ParallelWaveGAN}}

To train and test the TTS model, we use the data described in Section \ref{ccreation}, but only for a single speaker (Blizzard Challenge 2013; this speaker has read all 6 books). While the training corpus is read by a single speaker, this speaker portrays several different characters with different accents and pitch ranges. This tends to introduce fuzziness into the synthetic speech. To help the model learn these characteristics and improve quality, we added a speaker embedding to the TTS input. To obtain a speaker embedding, we 1) encoded each utterance in the training set with a pretrained speaker identification model (ECAPA-TDNN \cite{DBLP:conf/interspeech/DesplanquesTD20} available at \cite{speechbrain}), 2) used k-means clustering on these embeddings to obtain 30 different  `speakers' and 3) used these speaker labels as an additional input to FastSpeech 2; the speaker embeddings are concatenated with the phoneme and prominence label embeddings. 

\noindent \textbf{Listening Test}. To test the controllability of our TTS model in terms of prominence, we conducted an ordinal ranking listening test using the Web Audio Evaluation Tool \cite{waet2015}, following this procedure: 1) we randomly sampled 100 utterances containing pronouns from our full test set (not solely from the contrastive subset); 2) from this selection, we took the first 10 utterances containing a subjective pronoun, the first 10 with an objective pronoun and the first 10 with a possessive determiner (for a total of 30 utterances); 3) using our pretrained TTS system, we synthesized three versions of each utterance, changing only the prominence label for the relevant pronoun ($c\in \{<$p0$>$,$<$p1$>$,$<$p2$>$\}). The prominence labels for all other words in the sentence were kept constant with the ground truth values (extracted from the original audio with the CWT method); 4) 30 native English evaluators, recruited on Prolific,\footnote{\scriptsize \url{https://www.prolific.co}} were presented with the three versions of the synthesized utterances (in random presentation order) and a transcript of the audio with the pronoun of interest in uppercase letters. Participants were asked to rank the prominence of the pronoun by dragging and dropping the movable audio clips so that they were arranged from most prominent to least; 5) clips ranked as most prominent are assigned a score of 1; clips ranked as second most prominent are assigned a score of 0.5 and clips ranked as least prominent are given a score of 0. Sample audio files are available in the supplementary multimedia materials.

\begin{figure}[t]
  \centering
  \includegraphics[width=\linewidth]{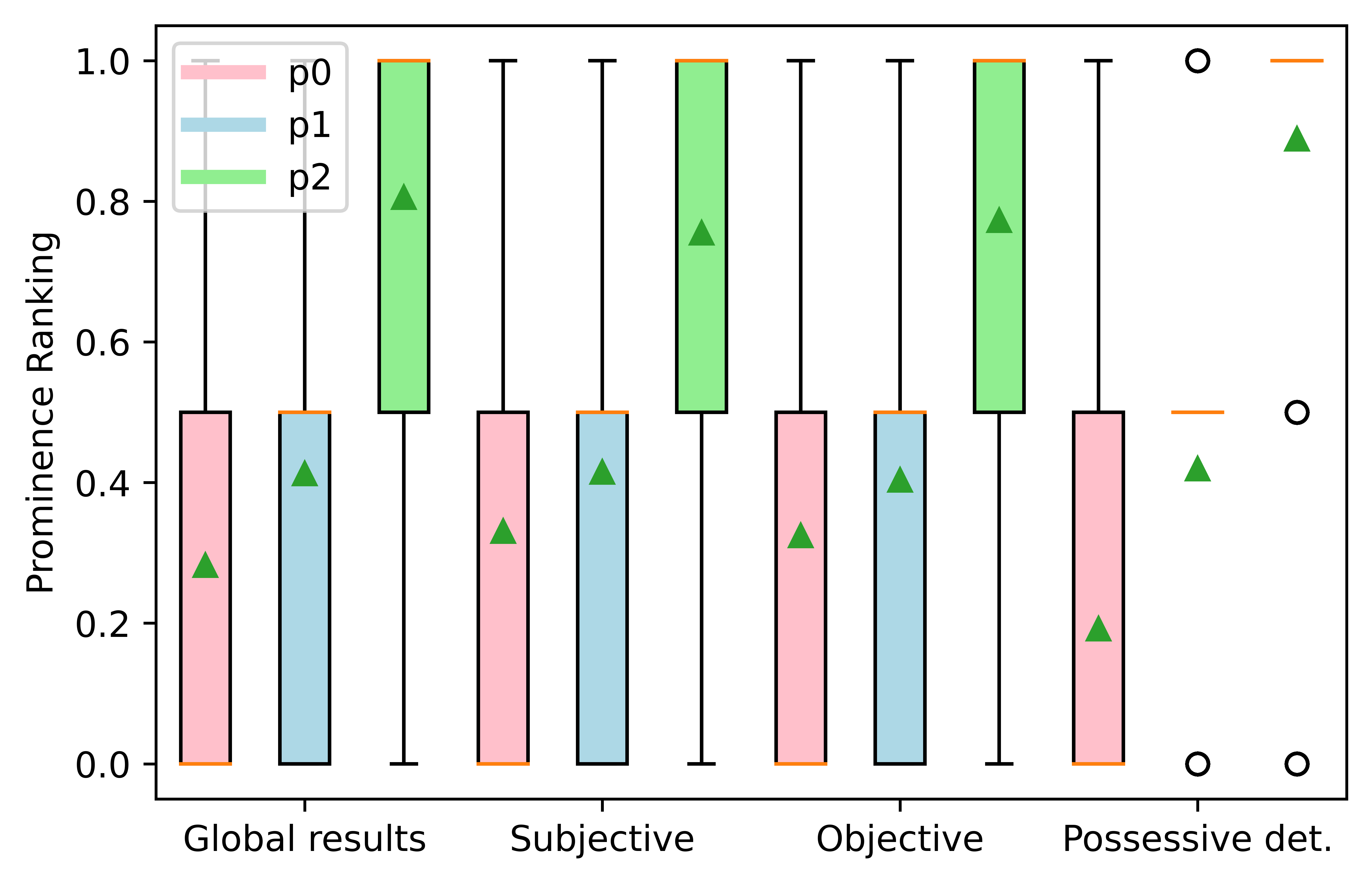}
  \caption{Results from the prominence ordinal ranking listening test. Audio ranked as most prominent was assigned a score of 1, second most prominent 0.5 and least prominent 0. Orange lines show the median and green triangles show the mean.}
  \label{fig:controllability}
\end{figure}

\noindent \textbf{Results}. The results of the listening test are shown in Figure~\ref{fig:controllability}. We see the median values align with the prosody labels used (median: $<$p0$>=0$, $<$p1$>=0.5$, $<$p2$>=1.0$). We do however see a wide distribution in the responses. This, and the examination of the ratings for individual utterances, indicates that this method works, but not consistently (i.e., there are some utterances for which the evaluators could not detect a difference). The only pronoun category for which we see a fairly clear distinction between $<$p0$>$, $<$p1$>$, and $<$p2$>$ in perceived prominence is for possessive determiners. This may be because there is more natural variation within the training corpus for this category. Or, it may be due to the labelling errors at phrase boundaries discussed in the previous section: the sampled subjective and objective pronouns were found more often at the beginning and end of phrases than the possessive determiners. More work is thus needed to disentangle the global prosodic representations from that of individual words, but this separation is difficult because it may result is less natural utterances.


\section{Conclusion and Perspectives}

In this paper, we investigated the difficult tasks of prominence prediction and control for contrastively focused personal pronouns. \textbf{Prediction:} we compared models with varying degrees of access to linguistic knowledge and past context on a word-level prominence label prediction task. We found that a finetuned BERT gave the best prediction performance, but that the improvement over baselines was very small. In the future, we will investigate the use of word representations from larger language models with more sophisticated linguistic understanding. \textbf{Control:} with a perceptive test, we evaluated the control of prominence in a TTS model conditioned on prominence labels. The results show the model is able to provide some control but the performance is not consistent over all samples.
\noindent \textbf{Natural variation:} we used multiple spoken versions of the same written text to see the agreement among speakers on the use of contrastive focus. But we must keep in mind that while the textual context remains the same, the interpretation of the text can vary. For example, we infer that one of the speakers interprets some of the characters in the novel to be passive aggressive and they convey this through the frequent use of contrastive focus on `I' (e.g., \textbf{I} am going to Gretna Green (intended meaning: and \textbf{YOU} are not). Removing the contrastive focus here is not wrong, but gives a very different impression of character/situation/relationship. This illustrates the difficulty of the prediction task and therefore, depending on the intended usage of the TTS system, it might be fruitful to explore other sources of input to the prominence prediction model (e.g., the source speech in a speech-to-speech translation system) in order to be as faithful to the intended meaning as possible.

\section{Acknowledgements}

Work was funded by the Multidisciplinary Institute in Artificial Intelligence MIAI@Grenoble-Alpes (ANR-19-P3IA-0003).

\bibliographystyle{IEEEtran}

\bibliography{mybib}


\end{document}